\documentclass[sigconf]{acmart}

\AtBeginDocument{%
  \providecommand\BibTeX{{%
    \normalfont B\kern-0.5em{\scshape i\kern-0.25em b}\kern-0.8em\TeX}}}

\copyrightyear{2023}
\acmYear{2023}
\setcopyright{acmcopyright}
\acmConference[CODS-COMAD 2023]{6th Joint International Conference on Data Science & Management of Data (10th ACM IKDD CODS and 28th COMAD)}{January 4--7, 2023}{Mumbai, India}
\acmBooktitle{6th Joint International Conference on Data Science \& Management of Data (10th ACM IKDD CODS and 28th COMAD) (CODS-COMAD 2023), January 4--7, 2023, Mumbai, India}
\acmPrice{15.00}
\acmDOI{10.1145/3570991.3571060}
\acmISBN{978-1-4503-9797-1/23/01}

\newcommand{\digitour}{\mathtt{DIGITOUR}}
\usepackage{amsmath}
\usepackage{enumitem,kantlipsum}

\usepackage{orcidlink}
\usepackage{graphicx}

\newcommand*{\img}[1]{%
    \raisebox{-.3\baselineskip}{%
        \includegraphics[
        height=\baselineskip,
        width=\baselineskip,
        keepaspectratio,
        ]{#1}%
    }%
}

\begin{document}

\title{$\digitour$: Automatic \textbf{Digi}tal \textbf{Tour}s for Real-Estate Properties}

\author{Prateek Chhikara}
\orcid{0000-0003-4833-474X}
\affiliation{
 \institution{Housing.com}
 \city{Gurgaon}
 \country{India}}
 
\email{prateek.chhikara@housing.com}

\author{Harshul Kuhar}
\affiliation{%
 \institution{Housing.com}
 \city{Gurgaon}
 \country{India}}
\email{harshul.kuhar@housing.com}

\author{Anil Goyal}
\affiliation{%
  \institution{Housing.com}
 \city{Gurgaon}
 \country{India}}
\email{anil.goyal@housing.com}

\author{Chirag Sharma}
\affiliation{%
 \institution{Housing.com}
 \city{Gurgaon}
 \country{India}}
\email{chirag.sharma@housing.com}

\renewcommand{\shortauthors}{P. Chhikara et al.}

\begin{abstract}
A virtual or digital tour is a form of virtual reality technology which allows a user to experience a specific location remotely. 
Currently, these virtual tours are created by following a $2$-step strategy. First, a photographer clicks a $360^\circ$ equirectangular image; then, a team of annotators manually links these images for the ``walkthrough'' user experience.
The major challenge in the mass adoption of virtual tours is the time and cost involved in manual annotation/linking of images.
Therefore, this paper presents an end-to-end pipeline to automate the generation of 3D virtual tours using equirectangular images for real-estate properties. 
We propose a novel HSV-based coloring scheme for paper tags that need to be placed at different locations before clicking the equirectangular images using $360^\circ$ cameras.
These tags have two characteristics: \textit{i)} they are numbered to help the photographer for placement of tags in sequence and; \textit{ii)} bi-colored, which allows better learning of tag detection (using YOLOv5 architecture) in an image and digit recognition (using custom MobileNet architecture) tasks.
Finally, we link/connect all the equirectangular images based on detected tags.
We show the efficiency of the proposed pipeline on a real-world equirectangular image dataset collected from the Housing.com database.

\end{abstract}

\ccsdesc[500]{Computing methodologies~Object detection}
\ccsdesc[500]{Computing methodologies~Object recognition}



\keywords{Computer Vision, Real-Estate, Virtual Tour, Digit Recognition.}

\maketitle

\section{Introduction}
Over the past few years, the demand for online real-estate tools has increased drastically due to the ease of accessibility to the Internet, especially in countries like India \cite{ullah2021modelling}. 
There are many online real-estate platforms (e.g., Housing.com, Proptiger.com, Makaan.com, etc.) for owners, developers, and real-estate brokers to post properties for buying and renting purposes. Daily, these platforms receive $8,000$ to $9,000$ new listings.
Till now, the users on these platforms view images, snapshots, or videos, which may not build desired confidence to make a decision and finalize the deal.
To overcome this challenge and to enhance user experience, virtual tours are a potential solution.
Virtual tours are images linked together, allowing viewers to experience a particular location remotely (a frequently used example is Google Street View \cite{anguelov2010google}).
Recently, the demand for virtual tours has increased as they provide better interaction with users/customers, especially in businesses like Real-Estate \cite{sulaiman2020matterport}, Hotels \& Restaurants \cite{mohammad2009development}, Universities \& Schools, etc.
Broadly, there are three categories of virtual tours: \textit{i)} 2D video tours, \textit{ii)} $360^\circ$ video-based virtual tours, and \textit{iii)} $360^\circ$ static image-based virtual tours. 
Compared to 2D video tours and $360^\circ$ video-based tours, the static equirectangular image-based virtual tours provide more immersion and interactivity, thus leading to better decision-making and avoiding unnecessary visits.
\begin{figure}[!htbp]
    \includegraphics[width=0.7\linewidth]{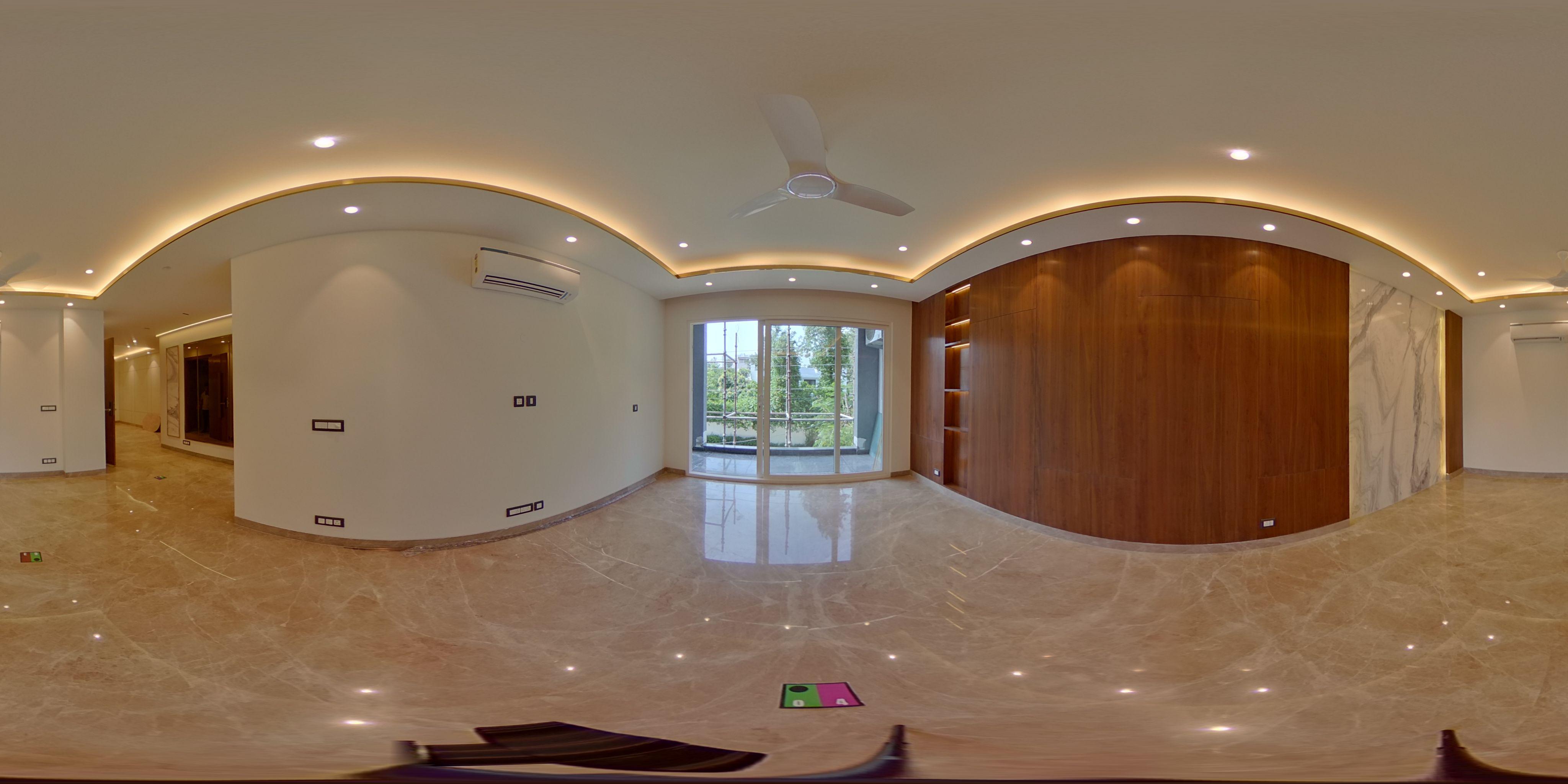}
    \caption{An example of Equirectangular image clicked at the entry of a house showing the living room.}
    \label{fig:equirectangular_image}
    \end{figure}

    \begin{figure*}[!htbp]
      \includegraphics[width=0.8\linewidth]{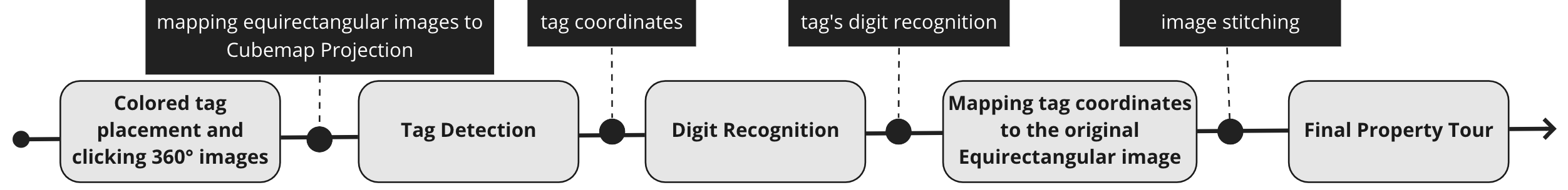}
      \caption{End-to-end pipeline for proposed approach $\digitour$}
      \label{fig:flow_diagram}
\end{figure*}
\begin{figure}[!htbp]
      \includegraphics[width=0.75\linewidth]{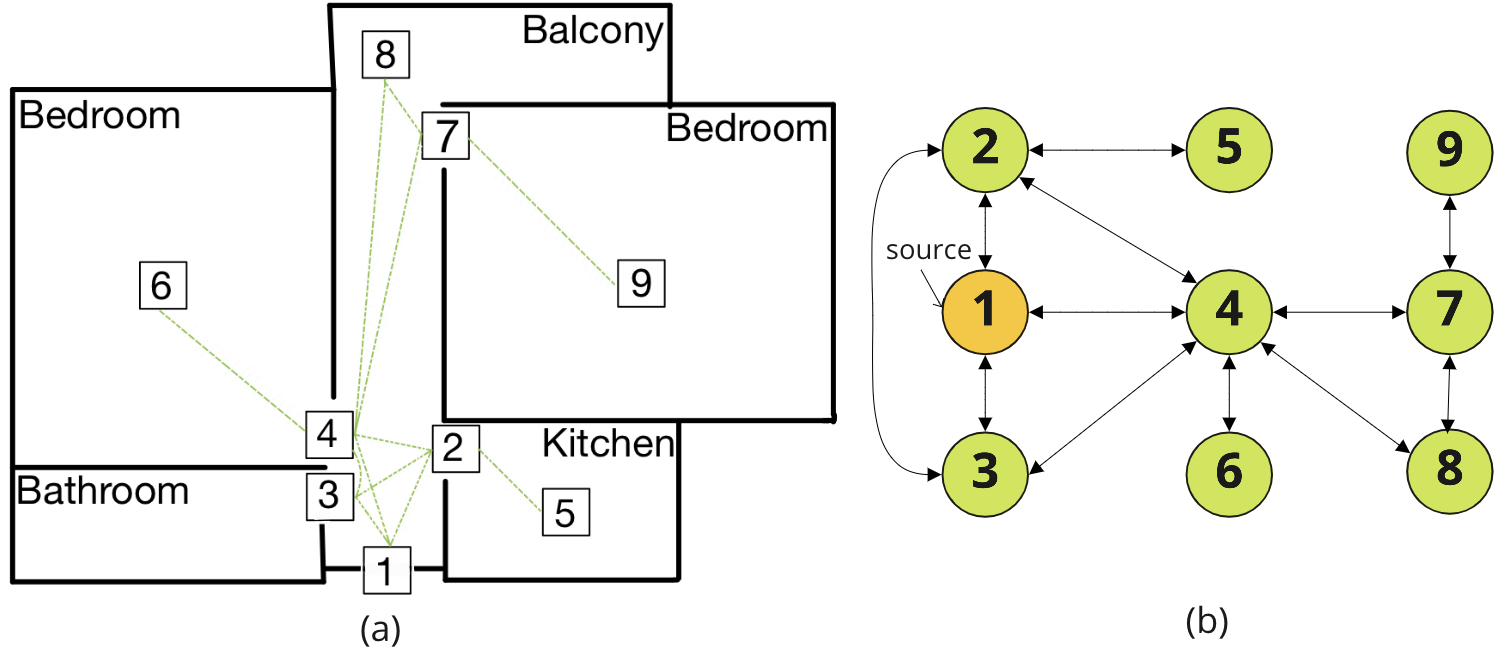}
      \caption{Example property floor plan (a) and it's connections (b) to make a digital tour. The numbers represent the position of tags and equirectangular images.}
      \label{fig:graph}
\end{figure}

Typically, the pipeline for creating a virtual tour consists of the following components:
\begin{enumerate}[leftmargin=*]
    \item[i)] \textbf{Equirectangular Image Capture:} An equirectangular image is a way of representing a spherical object as a 2D image (as shown in Figure \ref{fig:equirectangular_image}).  
    Basically, it is a spherical panorama incorporating a 180° vertical viewing angle and a 360° horizontal viewing angle. 
    A simple example can be the projection of Earth (a spherical-shaped object) on a 2D map. 
    These images are clicked using $360^\circ$ cameras such as Ricoh-Theta\footnote{\label{footnote:ricoh-theta}\url{https://theta360.com/en/etc/technology.html}}, Insta360\footnote{\url{https://www.insta360.com/product/insta360-onex2}}, etc.
    \item[ii)] \textbf{Connecting Equirectangular Images:} For any location, we will have multiple equirectangular images. To illustrate, in real-estate property, we typically have equilateral images for bedroom, hall, kitchen, dining room, etc. It is essential to build navigation between images to have a complete ``walkthrough'' experience. 
    Moreover, there can be multiple routes from one position to other positions. For instance, we can go from the hall to the kitchen, bedroom, balcony, etc. Therefore, it is crucial to connect all the equirectangular images (please refer to Figure \ref{fig:graph} for an example). 
    Generally, it is done manually, which is both costly and time-consuming \cite{mindtree}.
    \item[iii)] \textbf{Publishing Virtual Tour:} Once we have formed connections between equirectangular images, we can publish the final virtual tour on the cloud or in an application.
\end{enumerate}
  

While automating the above pipeline, one of the major challenges is the manual annotation for connecting equirectangular images.
Generally, it takes more than $20$ minutes to create a virtual tour. 
To overcome the challenge of creating automated \textbf{digi}tal \textbf{tour}s, we propose an end-to-end pipeline for real-estate equirectangular images (called $\digitour$).
The $\digitour$ pipeline consists of the following components:
\begin{enumerate}[leftmargin=*]
    \item[i)] \textbf{Colored tag placement and clicking 360° images:} We propose novel paper tags which are bi-colored and numbered, facilitating better learning of downstream computer vision tasks (i.e., tag recognition and digit recognition) and automatically stitching of equirectangular images.

    
    \item[ii)] \textbf{Map equirectangular images to cubemap projection:} We used publicly available python library \texttt{vrProjector}\footnote{https://github.com/bhautikj/vrProjector} to map equirectangular images to their cubemap projections (corresponding to six cube faces).
    \item[iii)] \textbf{Tag Detection:} For each cube face, we propose colored tag detection in an image using YOLOv5 \cite{glenn_jocher_2022_6222936} architecture.
    \item[iv)] \textbf{Digit Recognition:} We propose to perform digit recognition using a light-weight custom MobileNet model. \cite{sandler2018mobilenetv2}. 
\end{enumerate}

Finally, we connect all the equirectangular images using the detected tags. 
We experimentally validate the individual tasks (tag detection and digit recognition) and end-to-end pipeline on two datasets (``Green Colored Tags'' and ``Bi-colored Tags''). 
Our experiments show that the end-to-end pipeline performance is $88.12$ and $95.81$ in terms of mAP and f1-score at 0.5 IoU threshold averaged (weighted) over all the classes. 
In the next section, we present the proposed algorithm $\digitour$ followed by experimental results in Section \ref{sec:experiments}.

\begin{figure}[!ht]
    \centering
\includegraphics[width=0.2\linewidth]{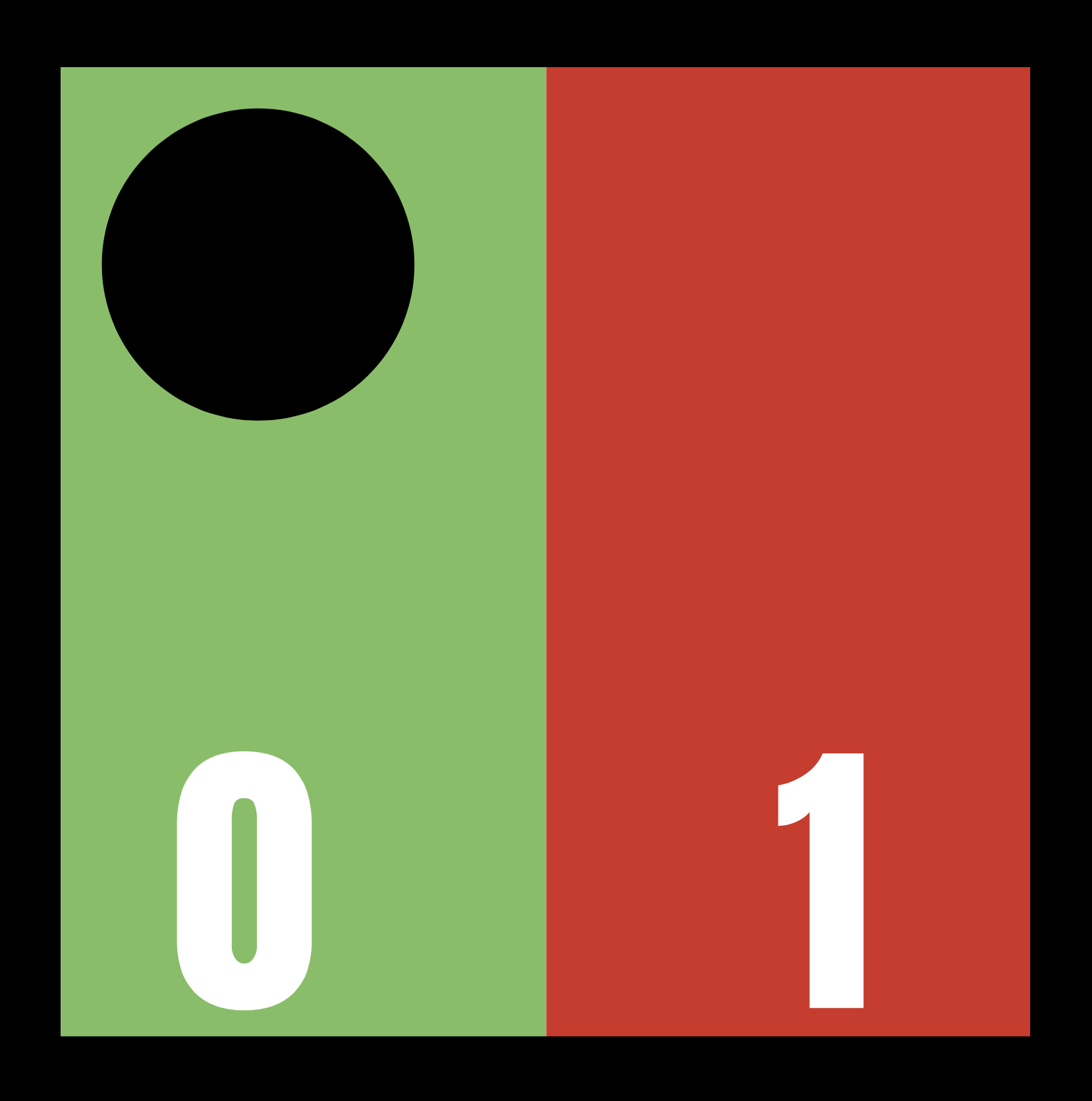}
\hspace{0.7cm}
\begin{tabular}{llc}
  \toprule
\textbf{Digit} & \textbf{Color}      & \textbf{HSV value} \\ \midrule
0     & \textcolor[HTML]{5fc24f}{Green}      &   (112°, 59\%, 76\%)        \\ \hline
1     & \textcolor[HTML]{d62b22}{Red}        &     (3°, 84\%, 84\%)      \\ \hline
2     & \textcolor[HTML]{7f3bd9}{Violet}     &    (266°, 73\%, 85\%)       \\ \hline
3     & \textcolor[HTML]{944d1b}{Brown}      &     (25°, 82\%, 58\%)   \\ \hline
4     & \textcolor[HTML]{de5bbb}{Pink}       &    (316°, 59\%, 87\%)       \\ \hline
5     & \textcolor[HTML]{a3a3a3}{Grey}       &    (0°, 0\%, 64\%)       \\ \hline
6     & \textcolor[HTML]{e3b540}{Yellow}     &    (43°, 72\%, 89\%)       \\ \hline
7     & \textcolor[HTML]{54bede}{Light Blue} &    (194°, 62\%, 87\%)       \\ \hline
8     & \textcolor[HTML]{0744ed}{Dark Blue}  &    (224°, 97\%, 93\%)       \\ \hline
9     & \textcolor[HTML]{f08132}{Orange}     &   (25°, 79\%, 94\%)        \\ \bottomrule
    \end{tabular}
    \captionlistentry[table]{A table beside a figure}
    \caption{Proposed bi-colored tag format and color scheme for each digit with their corresponding HSV values.}
    \label{fig:tag_image}
\end{figure}







\section{$\digitour$}
This section presents the complete pipeline for producing digital tours (referred to as $\digitour$ and shown in Figure \ref{fig:flow_diagram}).

\subsection{Tag Placement and Image Capturing}
While creating a digital tour for any real-estate property, it is essential to click $360^\circ$ images from different property locations such as bedroom, living room, kitchen, etc.,
then automatically stitching them together to have a ``walkthrough'' experience without being physically present at the location.
Therefore, to connect multiple equirectangular images, we propose placing paper tags 
on the floor covering each location of the property, and placing the camera (in our case, we used Ricoh-Theta) in the middle of the scene to capture the whole site (front, back, left, right and bottom). Moreover, we ensure that the scene is clear of all noisy elements such as dim lighting and `unwanted' artifacts for better model training and inference. 
As shown in Figure \ref{fig:tag_image}, we have standardized the tags with dimensions of $6" \times 6"$ with two properties: \textit{i)} they are numbered which will help the photographer place tags in sequence and \textit{ii)} they are bi-colored to formulate the digit recognition problem as classification task and facilitate better learning of downstream computer vision tasks (i.e. tag detection and digit recognition).
Please note that different colors are assigned to each digit (from $0$ to $9$) using the HSV color scheme and leading digit of a tag has a black circle to distinguish it from the trailing digit as shown in Figure \ref{fig:tag_image}. 
The intuition behind standardizing the paper tags is that it allows to train tag detection and digit recognition models, which are invariant to distortions, tag placement angle, reflection from lighting sources, blur conditions, and camera quality.

\begin{figure}[!htbp]
  \includegraphics[width=0.7\linewidth]{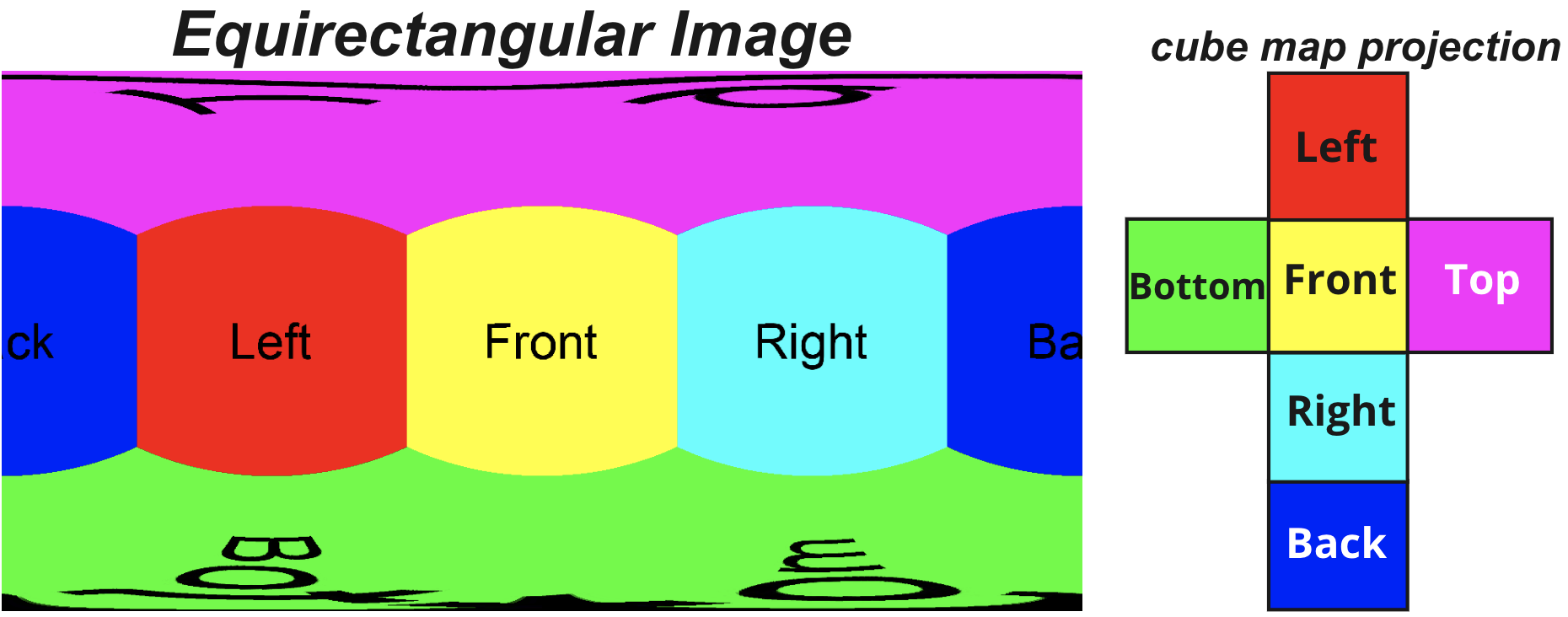}
  \caption{Conversion of an equirectangular image to its corresponding six faces cubemap projection.}
  \label{fig:step1}
\end{figure}


\vspace{-10pt}
\subsection{Mapping Equirectangular Image to Cubemap Projection}

An equirectangular image consists of a single image whose width and height correlate as $2:1$ (as shown in Figure \ref{fig:equirectangular_image}). 
In our case, images are clicked using a Ricoh-Theta camera having dimensions  $4096 \times 2048 \times 3$. 
Typically, each point in an equirectangular image corresponds to a point in a sphere, and the images are stretched in the `latitude' direction. 
Since the contents of an equirectangular image are distorted, it becomes challenging to detect tags and recognize digits directly from it. 
For example, in Figure \ref{fig:equirectangular_image}, the tag is stretched at the middle-bottom of the image. 
Therefore, it is necessary to map the image to a less-distorted projection and switch back to the original equirectangular image to build the digital tour. 
In this work, we propose to use cubemap projection, which is a set of six images representing six faces of a cube.
Here, every point in the spherical coordinate space corresponds to a point in the face of the cube. 
As shown in Figure \ref{fig:step1}, we map the equirectangular image to six faces (left, right, front, back, top and bottom) of a cube having dimensions  $1024\times1024\times3$ using python library \texttt{vrProjector}.


\subsection{Tag Detection}
\label{sec:tag_detection}
\begin{figure}[!htbp]
  \includegraphics[width=0.7\linewidth]{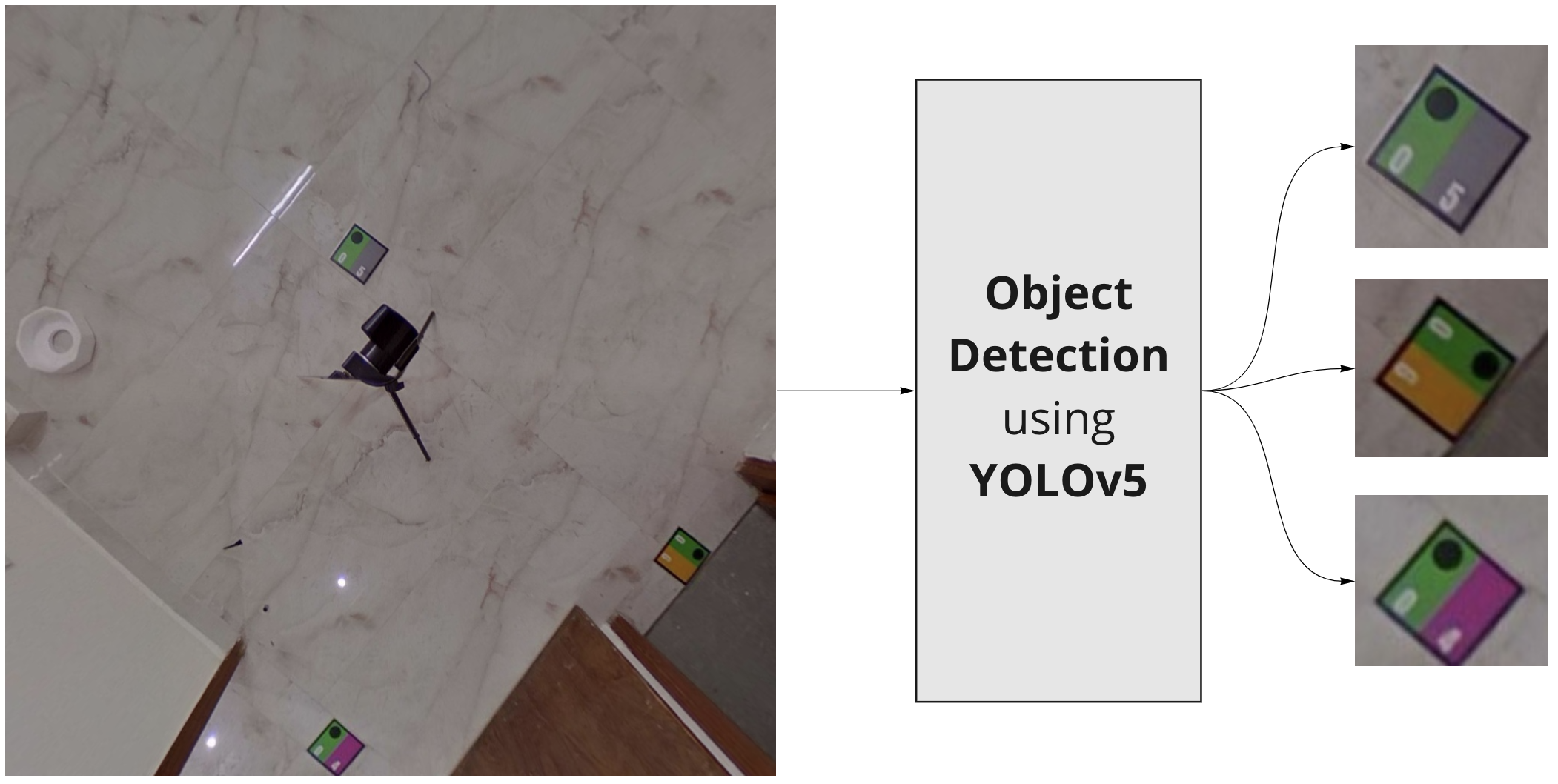}
  \caption{Tag detection using Yolov5.}
  \label{fig:tag_detection}
\end{figure}

Once we get the six images corresponding to the faces of a cube, we detect the location of tags placed in each image. 
For tag detection, we have used the state-of-the-art YOLOv5 \cite{glenn_jocher_2022_6222936} model. We initialized the network with COCO weights followed by training on our dataset. 
As shown in Figure \ref{fig:tag_detection}, the model takes an image as input and returns the detected tag along with coordinates of the bounding box and confidence of the prediction. 
The model is trained on our dataset for $100$ epochs with a batch size of 32.


\subsection{Digit Recognition}
\label{sec:digit_recognition}
\begin{figure}[!htbp]
  \includegraphics[width=0.7\linewidth]{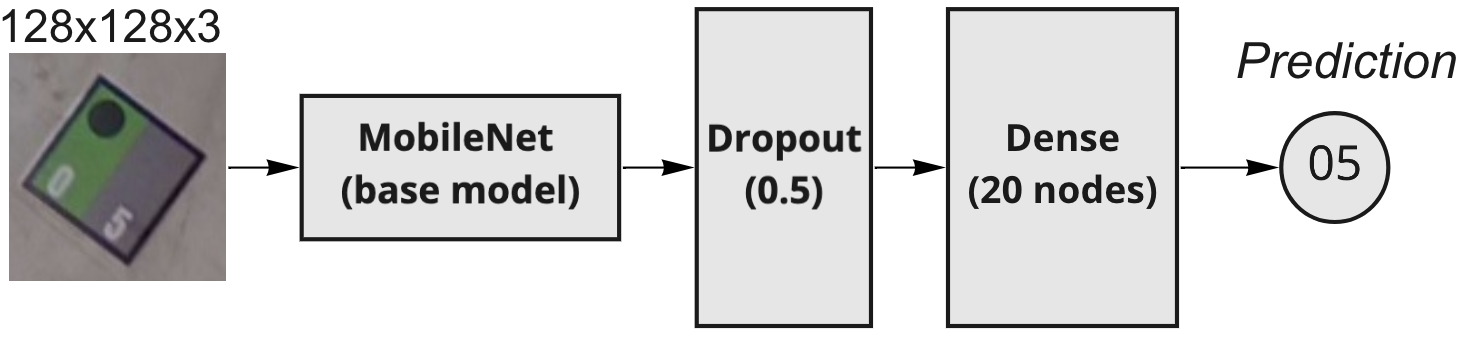}
  \caption{Digit recognition using custom MobileNet model.}
  \label{fig:digit_recognition}
\end{figure}

For the detected tags, we need to recognize the digits from the tag. 
In a real-world environment, the detected tags might have incorrect orientation, poor luminosity, reflection from the bulbs in the room, etc. 
Due to these reasons, it is challenging to use Optical Character Recognition (OCR) engines to have good digit recognition performance. 
Therefore, we have used a custom MobileNet \cite{sandler2018mobilenetv2} model initialized on imagenet weights, which uses color information in tags for digit recognition. 
In the proposed architecture, we have replaced the final classification block of the original MobileNet with the dropout layer and dense layer with 20 nodes representing our tags from 1 to 20. Figure \ref{fig:digit_recognition} illustrates the proposed architecture.
For training the model, we have used Adam as an optimizer
with a learning rate of 0.001 and a discounting factor ($\rho$) to be 0.1. We have used categorical cross-entropy as a loss function and set the batch size to 64 and the number of epochs to 50.

\vspace{-15pt}
\subsection{Mapping tag coordinates to the original $360^\circ$ Image and Virutal Tour Creation}
\begin{figure}[!htbp]
  \includegraphics[width=0.5\linewidth]{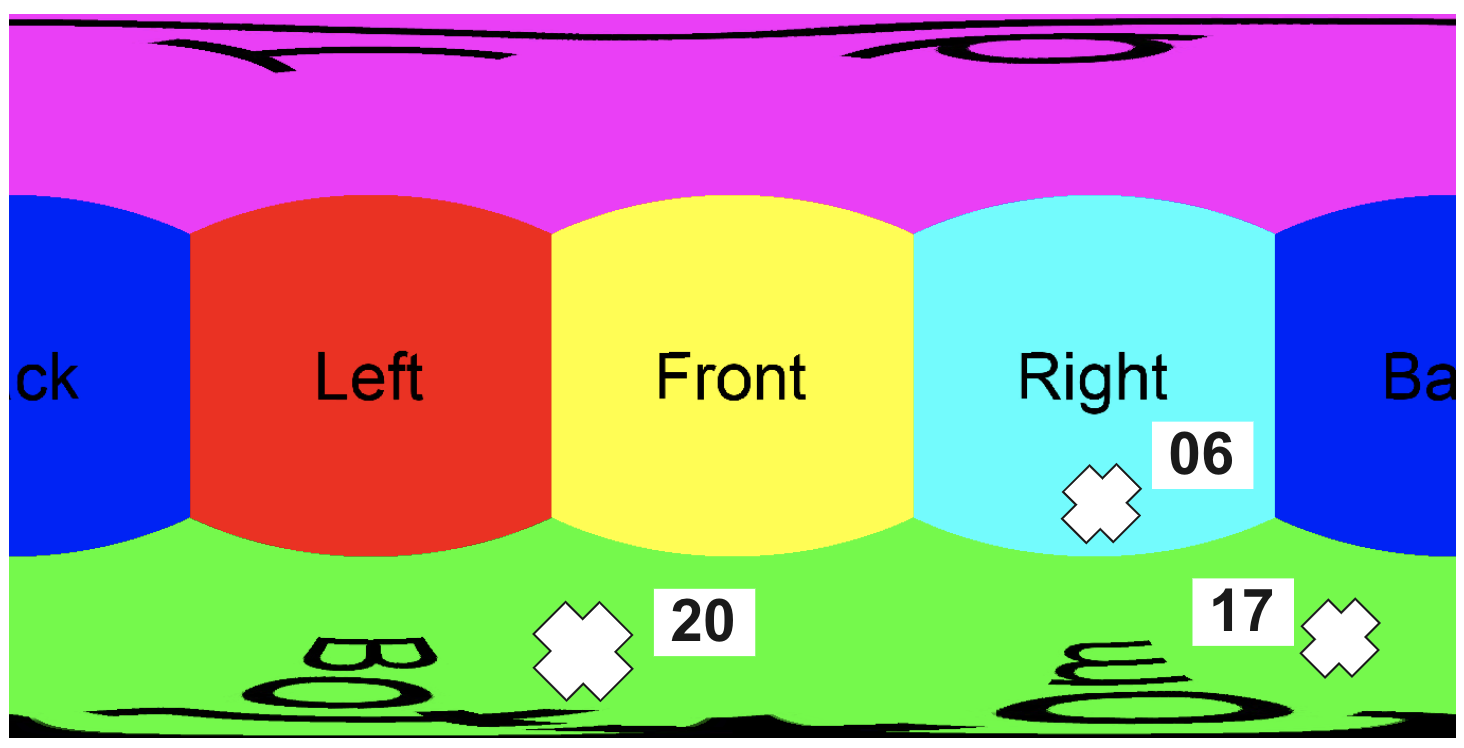}
  \caption{Mapping tags to original equirectangular image.}
  \label{fig:map_coordinates}
\end{figure}

Once we have detected the tags and recognized the digits we use the python library $\mathtt{vrProjector}$ to map the cubemap coordinates back to the original equirectangular image. An example output is shown in Figure \ref{fig:map_coordinates}.
For each equirectangular image, the detected tags form the nodes of a graph with an edge
between them. In the subsequent equirectangular images of a property, the graph gets populated with more nodes, as more tags are detected. 
Finally, we connect multiple equirectangular images in sequence based on recognized digits written on them and the resulting graph is the virtual tour as shown in Figure \ref{fig:graph}(b).

\section{Experiments}
\label{sec:experiments}
In this section, we present an empirical study to show the performance of our pipeline $\digitour$.

\subsection{Datasets}
We have collected data by placing tags and clicking equirectangular images using Ricoh-Theta camera for several residential properties in Gurugram, India (Tier 1 city).
While collecting images we made sure that certain conditions were met such as all doors were opened, lights were turned on, `unwanted' objects were removed and the tags were placed covering each area of the property.
Following these instructions, average number of equirectangular images clicked per residential property was $7$ or $8$.
Finally, we have validated our approach on the following generated datasets (based on background color of the tags). 
\begin{itemize}[leftmargin=*]
    \item \textbf{Green Colored Tags} (\img{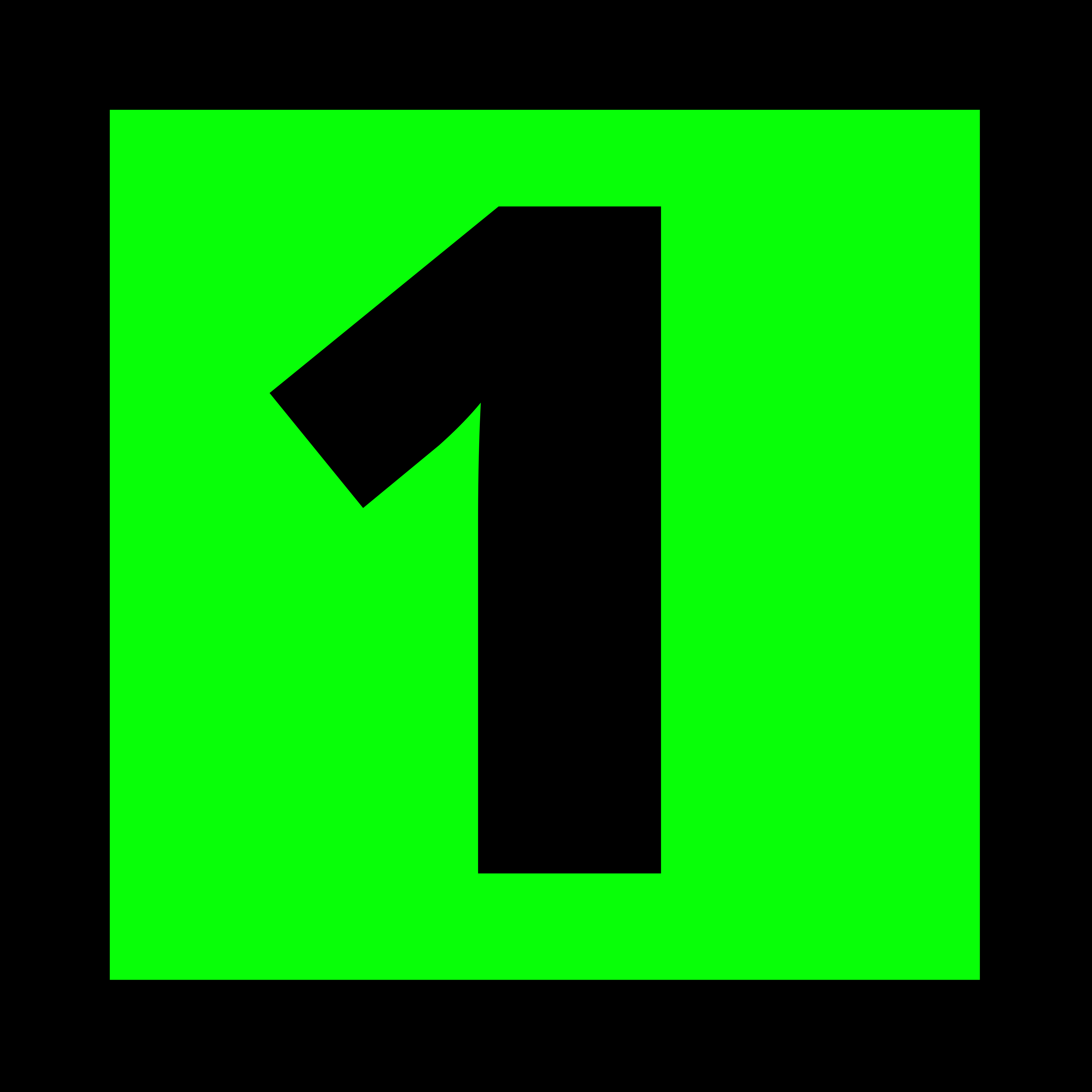}): We have kept the background color of these tags (numbered $1$ to $20$) to be green. 
    We have collected $1572$ equirectangular images from $212$ properties.
    Once we convert these equirectangular images to cubemap projection, we get $9432$ images (corresponding to cube faces).
    Since not all of the cube faces have tags (for e.g. top face), we get $1503$ images with atleast one tag. 
    \item \textbf{Proposed Bi-colored Tags} (\img{figures/tag.png}):
    For these tags, we have collected $2654$ equirectangular images from 350 properties.
    Finally, we got $2896$ images (corresponding to cube faces) with atleast one tag.
\end{itemize}
Finally, we label the tags present in cubemap projection images using LabelImg\footnote{https://github.com/tzutalin/labelImg} which is an open-source tool for labeling images in several formats such as Pascal VOC and YOLO.
For all the experiments, we reserved 20\% of data for testing and the remaining for training.


\subsection{Results}
\subsubsection{Evaluation Metrics} 
As followed in literature \cite{padilla2020survey}, we used accuracy (A), precision (P), recall (R), f1-score (f1) and mean average precision (mAP@k) for the tag detection task at different intersection over union (IoU) thresholds (represented by k). 
We treat digit recognition as a classification task, therefore we used macro and weighted average accuracy (A), precision (P), recall (R) and f1-score (f1) over all the classes as evaluation metrics \cite{grandini2020metrics}. 


\begin{table}[!htbp]
\caption{Tag detection performance (\%)}
\label{table:tag_detection}
\begin{tabular}{lcccc}
\toprule
\textbf{Tag Type} & \textbf{P} & \textbf{R} & \textbf{mAP @0.5} & \textbf{mAP @0.95} \\ \midrule
\textbf{Green colored} & 96.5 & 98.2 & 99.2 & 67.7 \\ \hline
\textbf{Bi-colored}  & 98.5 & 98.4 & 99.3 & 71.4 \\ \bottomrule
\end{tabular}
\end{table}

\subsubsection{Tag Detection}
We have trained YOLOv5 models for both datasets keeping the same experimental setting as discussed in Section \ref{sec:tag_detection}. 
From Table \ref{table:tag_detection}, we can deduce that the performance of YOLOv5 architecture for both of the colored tags is similar. There is no significant gain in performance with change in background color of tags. 


\subsubsection{Digit Recognition}
We have formulated the digit recognition task in two ways: \textit{i)} as OCR (optical character recognition) task and; \textit{ii)} as classification problem. 
On ``Green Colored Tag'' dataset, we experimentally validated PaddleOCR \cite{du2021pp} model and custom-trained  InceptionV3 \cite{szegedy2016rethinking} model. 
However, the obtained f1-scores for PaddleOCR and InceptionV3 models were $33\%$ and $24\%$. 
Finally, we experimentally validated various state-of-the-art models for digit recognition task on ``Bi-colored Tags'' dataset (as shown in Table \ref{digit_recog_table}). 
From Table \ref{digit_recog_table}, we can deduce that InceptionV3 \cite{szegedy2016rethinking} model performs best as compared to MobileNet \cite{sandler2018mobilenetv2}, VGG19 \cite{simonyan2014very} and ResNet50 \cite{he2016deep}. 
Moreover, InceptionV3 and MobileNet architectures have similar performance on the bi-colored tag dataset.
However, the average response time per image for InceptionV3 and MobileNet  model is 8.26 ms and 2.74 ms respectively.
Therefore, we decided to choose MobileNet as final model for digit recognition task. 


\begin{table}[!htbp]
\caption{Comparison of different state-of-the-art models on bi-colored tags dataset for digit recognition task.}
\label{digit_recog_table}
\begin{tabular}{lrrrrrrr}
\toprule
\multicolumn{2}{l}{\textbf{}} & \multicolumn{3}{c}{\textbf{Macro}} & \multicolumn{3}{c}{\textbf{Weighted Avg.}} \\ \hline
\multicolumn{1}{l}{\textbf{Models}} & \multicolumn{1}{c}{\textbf{A}} & \multicolumn{1}{c}{\textbf{P}} & \multicolumn{1}{c}{\textbf{R}} & \multicolumn{1}{c}{\textbf{f1}} & \multicolumn{1}{c}{\textbf{P}} & \multicolumn{1}{c}{\textbf{R}} & \multicolumn{1}{c}{\textbf{f1}} \\ \midrule
\multicolumn{1}{l}{InceptionV3} & 99.2 & \multicolumn{1}{r}{98.1} & \multicolumn{1}{r}{98.6} & 98.2 & \multicolumn{1}{r}{99.2} & \multicolumn{1}{r}{99.3} & 99.3 \\ \hline
\multicolumn{1}{l}{MobileNet } & 98.9 & \multicolumn{1}{r}{99.0} & \multicolumn{1}{r}{98.6} & 98.8 & \multicolumn{1}{r}{99.0} & \multicolumn{1}{r}{98.9} & 98.9 \\ \hline
\multicolumn{1}{l}{VGG19 } & 96.1 & \multicolumn{1}{r}{96.1} & \multicolumn{1}{r}{94.5} & 95.2 & \multicolumn{1}{r}{96.2} & \multicolumn{1}{r}{96.1} & 96.1 \\ \hline
\multicolumn{1}{l}{ResNet50} & 93.4 & \multicolumn{1}{r}{90.8} & \multicolumn{1}{r}{92.8} & 90.5 & \multicolumn{1}{r}{94.3} & \multicolumn{1}{r}{93.4} & 93.5 \\ 
\bottomrule
\end{tabular}
\end{table}


\subsubsection{End-to-end Evaluation}
In this section, we present the experimental results for the proposed end-to-end pipeline. 
For any input image, we first detect the tags and finally recognize the digits written on the tags.
From this we were able to identify the true positives (tags detected and read correctly), false positives (tags detected but read incorrectly) and false negatives (tags not detected).
The obtained mAP, Precision, Recall and f1-score at 0.5 IoU threshold  are $88.12$, $93.83$, $97.89$ and $95.81$ respectively.
Please note that all metrics are averaged (weighted) over all the $20$ classes.
If all tags across all equirectangular images of a property are detected and read correctly, we receive a $100\%$ accurate virtual tour since all nodes of the graph are detected and connected with their appropriate edges.
 In our experiments, we were able to accurately generate $100\%$ accurate  virtual tour for $94.55\%$ of the properties.
The inaccuracies were due to the presence of colorful artifacts that were falsely detected as tags; and bad lightning conditions.


\section{Conclusion}
We propose an end-to-end pipeline ($\digitour$) for automatically generating digital tours for real-estate properties.
For any such property, we first place the proposed bi-colored paper tags covering each area of the property.
Then, we click equirectangular images, followed by mapping these images to less distorted cubemap images. 
Once we get the six images corresponding to cube faces, we detect the location of tags using the YOLOv5 model, followed by digit recognition using the MobileNet model. 
The next step is to map the detected coordinates along with recognized digits to the original equirectangular images.
Finally, we stitch together all the equirectangular images to build a virtual tour. 
We have validated our pipeline on a real-world dataset and shown that the end-to-end pipeline performance is $88.12$ and $95.81$ in terms of mAP and f1-score at 0.5 IoU threshold averaged (weighted) over all classes.

\bibliographystyle{ACM-Reference-Format}
\bibliography{main.bib}

\end{document}